\documentclass[a4paper,conference]{IEEEtran}
\usepackage{graphicx}
\usepackage{comment}
\usepackage{amsmath,amssymb} % define this before the line numbering.
\usepackage{color}
\usepackage{booktabs}
\usepackage{stfloats}

\usepackage[citecolor=green]{hyperref}
\newcommand{\blu}[1]{{\emph{\color{blue} {#1}}}}
\newcommand{\tbf}[1]{{\textbf{#1}}}

\ifCLASSINFOpdf
\else
\fi
\begin{document}
\title{Transformer Networks for Trajectory Forecasting}


% \author{\IEEEauthorblockN{Francesco Giuliari}
% \IEEEauthorblockA{University of Verona, Italy\\
% francesco.giuliari@univr.it}
% \and
% \IEEEauthorblockN{Irtiza Hasan}
% \IEEEauthorblockA{Inception Institute of Artificial Intelligence\\
% irtiza.hasan@inceptioniai.org}
% \and
% \IEEEauthorblockN{Marco Cristani}
% \IEEEauthorblockA{University of Verona, Italy\\
% marco.cristani@univr.it}
% \and
% \IEEEauthorblockN{Fabio Galasso}
% \IEEEauthorblockA{Sapienza University of Rome, Italy\\
% galasso@di.uniroma1.it}}

\author{\IEEEauthorblockN{Francesco Giuliari}
\IEEEauthorblockA{University of Verona\\
}

\and
\IEEEauthorblockN{Irtiza Hasan}
\IEEEauthorblockA{Inception Institute of Artificial Intelligence\\
}
\and
\IEEEauthorblockN{Marco Cristani}
\IEEEauthorblockA{University of Verona\\
}
\and
\IEEEauthorblockN{Fabio Galasso}
\IEEEauthorblockA{Sapienza University of Rome\\
}}

\maketitle



% \fi
\IEEEpeerreviewmaketitle
%
%Forecasting the people motion is an important but complex task, which has been addressed so far with a variety of temporal and spatio-temporal models. 
%
\begin{abstract}

Most recent successes on forecasting the people motion are based on LSTM models and \emph{all} most recent progress has been achieved by modelling the social interaction among people and the people interaction with the scene.
%nearly saturating a reference dataset ETH+UCY.
%
We question the use of the LSTM models and propose the novel use of Transformer Networks for trajectory forecasting. This is a fundamental switch from the sequential step-by-step processing of LSTMs to the only-attention-based memory mechanisms of Transformers. In particular, we consider both the original Transformer Network (TF) and the larger Bidirectional Transformer (BERT), state-of-the-art on all natural language processing tasks.
Our proposed Transformers predict the trajectories of the individual people in the scene. These are ``simple'' models because each person is modelled separately without any complex human-human nor scene interaction terms. In particular, the TF model \emph{without bells and whistles} yields the best score on the largest and most challenging trajectory forecasting benchmark of TrajNet~\cite{sadeghiankosaraju2018trajnet}. Additionally, its extension which predicts multiple plausible future trajectories performs on par with more engineered techniques on the 5 datasets of ETH~\cite{pellegrini2009iccv}+UCY~\cite{lerner2007crowds}. Finally, we show that Transformers may deal with missing observations, as it may be the case with real sensor data.
Code is available at \texttt{github.com/FGiuliari/Trajectory-Transformer}.

\end{abstract}

\section{Introduction}\label{sec:intro}

\begin{figure*}[t]
\centering
\includegraphics[width=\textwidth]{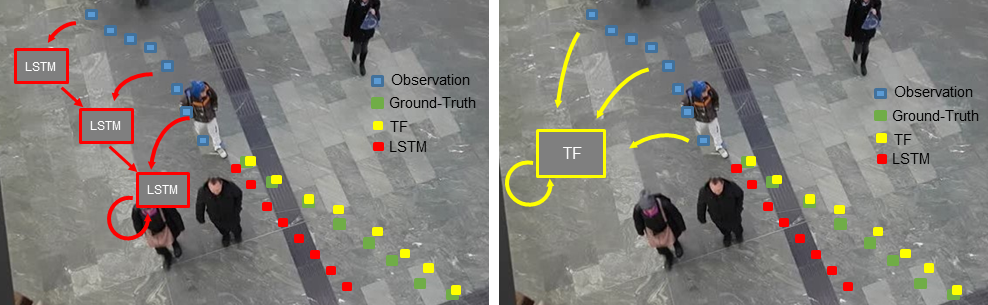}
\caption{People trajectory forecasting stands for predicting the future motion of people (\textit{green ground-truth dots}), given an observation interval (\textit{blue dots}). LSTM (left) sequentially processes the observations before starting to predict, while TF analyses in one shot all available observations.%Here we challenge the established LSTM approaches, at the base for \textit{all} state-of-the-art methods, by introducing Transformer Networks (TF), a revolutionary technique inspired by NLP most recent progress. 
}
\label{fig:teaser}
%\vspace{-0.3cm}
\end{figure*}

Pedestrian forecasting, the goal of predicting future people motion given their past trajectories, has been steadily growing in attention by the research community. Further to being a crucial compound of trackers, especially for the cases of large motion and/or missing observations, the topic serves early action recognition, surveillance and automotive systems. %Also importantly, pedestrian forecasting may support in avoiding collision between the autonomous cars and the people and in the human-robot interaction, as it provides the robots with the near-future human proxemics.

Starting from \cite{alahi2016cvpr}, Long Short-Term Memory (LSTM) networks have been the workhorse for forecasting and progress has been achieved by devising social pooling mechanisms to model the people social interaction~\cite{alahi2016cvpr,gupta2018social}. The LSTM is based on sequentially processing sequences and storing hidden states to represent knowledge about the people, e.g.\ its speed, direction and motion pattern. Most modern approaches have challenged each other on the social interaction of pedestrians, each modelled with a separate LSTM and exchanging information by means of social pooling mechanisms~\cite{alahi2016cvpr,gupta2018social}. In fact best performing approaches additionally include the semantics of the scene into the LSTMs~\cite{kosaraju2019social,salzmann2020trajectron++,ivanovic2019trajectron,sadeghian2019sophie}. However LSTMs have also been target of criticism: their memory mechanism has been criticised~\cite{bai2018empirical,Luo18} and, most recently, also their capability of modelling social interaction~\cite{scholler2020constant,becker2018red,becker2018evaluation}. An in-depth understanding of such mechanisms has not been supported by the adopted datasets, such as the 5 datasets of ETH~\cite{pellegrini2009iccv} and UCY~\cite{lerner2007crowds}, where performance measures are close to saturation, since leading techniques only report average forecasting errors of $\sim$20cm across 200m-long pavements. %\todo{refer to teaser. MARCO:WHY???}
%LSTM are the workhorse for forecasting, but they have inherent limitations (lack of long-term memory)---> opening experiment showing exponential degradation of performances, better with a social GAN model. Long time memory is important to model long term interactions, i.e., free social o-formations

In this work we side-step social and map mechanisms, and propose to model the trajectories of individual people by Transformer Networks~\cite{TransformersNIPS17}, for the first time. Transformer networks have been proposed for Natural Language Processing to model word sequences. These use attention instead of sequential processing. In other words, these estimate which part of the input sentence to focus on, when needing to translate, answer a question or complete the sentence~\cite{BERT19,SparseTF}. Here we consider for trajectory forecasting the original Transformer Network (TF) and the Bidirectional Transformer (BERT) models, on which state-of-the-art NLP algorithms are based. Fig.~\ref{fig:teaser} illustrates the fundamental difference between TF and LSTM: LSTM sequentially processes the observations before starting to predict auto-regressively, while TF ``looks'' at all available observations, weighting them according to an attention mechanism.

% A careful adaptation of TF and BERT to people forecasting yields results comparable to the state-of-the-art, without the need for modelling the social context. In particular, we find that data augmentation plays a role in adapting the models to dealing with datasets such as ETH+UCY, which are fairly smaller than NLP ones. Additionally, by moving from forecasting as a regression problem, as commonly addressed in literature, to a classification into quantized steps, we provide TF and BERT models capable of predicting multiple future outcomes.
%transformer has been applied in many NLP related contexts, and image processing, --> is single-agent native and should be revisited for trajectory forecasting, which naturally exploit contextual social knowledge

We assess the performance of TF and BERT on the TrajNet benchmark~\cite{sadeghiankosaraju2018trajnet}, in order to have a clean evaluation (TrajNet uses a unified evaluation system with a dedicated server) against 42 forecasting approaches, on a large selection of datasets.
Our TF outperforms all other techniques, also those including social mechanisms. TF compares favorably also on the ETH+UCY datasets, in particular beating all of the approaches that consider the individual trajectories only. %,  where we additionally test its capability to predict multiple plausible future outcomes. 
Finally we conduct an ablation to highlight the potential of the Transformers, quantitatively and qualitatively. Of particular interest is the ability of TF to still predict from inputs with missing observation data, thanks to its attention mechanism, which the LSTM cannot do.

%Our main contributions are:
%\begin{itemize}
%    \item We are the first to adopt Transformer Networks from NLP to trajectory forecasting;
%    \item we achieve state-of-the-art performance on the most challenging TrajNet benchmark \emph{without bells and whistle}, i.e.\ by only modelling the motion of person, without any social nor map-related term;
%    \item we achieve state-of-the-art performance on the 5 datasets of ETH+UCY, where we showcase a TF model which makes predictions of multiple plausible futures for each person's motion.
%\end{itemize}

%Among the merits of the proposed Social Transfomer architecture is a more suitable encoding of time. LSTM treats time as a quantized variable, it requires that any input be provided and it requires equally spaced temporal inputs, i.e.\ time steps. By contrast, our Social Transformer encodes the time-stamp of each input with a temporal encoding. This means that our Social Transformer may also deal with incomplete and sparse input sequences, where time steps need no be of a fixed time gap.

\vspace{-0.2cm}

\section{Related work}
\label{sec:prev}

% \todo{Irtiza: the current writing does not yet point out related work in a way which is suitable for the paper. Target:\\
% First part on traj forecasting:\\
% - classic approaches\\
% - lstm and rnn\\
% - big emphasis on social\\
% - most recently empahsis on even more, eg maps\\
% - by contrast we just consider a different sequence model, without bell and whistle, just individual\\
% Second part on transformers:\\
% - transformer\\
% - big in nlp\\
% - we introduce it in traj forecasting
% }

% Trajectory forecasting has historically witnessed two main breakthroughs: the introduction of social interaction and the modern era of data-driven modelling. The first has considered how people navigate in crowds~\cite{}, the second has unlocked the potential of convolutional and recurrent neural network techniques~\cite{}\todo{indicate first or representative works here}. Our work brings together teachings from both trends \todo{what do we take from crowds? TBC}, which we review here. Furthermore we review literature on Transformer networks, which we adopt for trajectory forecasting for the first time.

Forecasting people trajectories has been studied for over two decades and relevant literature has been surveyed by the work of \cite{becker2018evaluation,morris2008survey}.
For the purpose of this paper, we distinguish two main trends of related work: a first which has focused on progressing sequence modelling and a second which has modelled the interactions between the people and between the people and the scene.

\noindent\textit{Sequence modelling:}
Trajectory forecasting has experienced a steady progress from hand-crafted energy-based optimization approaches to data-driven ones.
Early work on human path prediction have adopted linear~\cite{mccullagh1989generalized} or Gaussian regression models ~\cite{quinonero2005unifying,williams1998prediction}, time-series analysis~\cite{priestley1981spectral} and auto-regressive models~\cite{akaike1969fitting}, optimizing for hand-crafted energy functions. By contrast, later models have been most successful by the adoption of LSTM~\cite{hochreiter1997long} and RNN models, trained with copious amounts of data.
In particular, LSTM can be employed to regress directly the predicted values~\cite{becker2018red,gupta2018social,salzmann2020trajectron++}, or to produce mean and (diagonal) covariance over the $x,y$ coordinates in order to express the uncertainty associated to the prediction~\cite{alahi2016cvpr}. In the latter case, we refer to this model as \emph{Gaussian LSTM}.
Here we argue that Transformer Networks are most suitable to sequence modelling and to forecast trajectories, thanks to their better capability to learn non-linear patterns, especially emerging when large amounts of data is available.

\noindent\textit{Social models and context:}
Enabled by the flexibility of the LSTM machinery, best performance has been recently achieved by modelling the social interaction~\cite{alahi2016cvpr,gupta2018social,vemula2018social} among people and the scene context~\cite{sadeghian2019sophie,salzmann2020trajectron++,kosaraju2019social}, aided by tracking dynamics~\cite{sadeghian2017tracking} and the spatio-temporal relations among neighboring people~\cite{ijcai2017-386,su2016crowd}.
Much literature has recently criticised the capability of LSTM to model the human-human interaction~\cite{scholler2020constant,becker2018red,becker2018evaluation}, maintaining that this limits the model generalization capability~\cite{scholler2020constant}. Our work side-steps social and environmental interactions and focuses on the prediction of the motion of each person individually. Somehow surprisingly, our ``simple'' approach achieves best performance on the most challenging benchmark of TrajNet.

In this work, we leverage findings and state-of-the-art techniques developed within the NLP field to model word sequences. In particular, we consider here for trajectory forecasting the original Transformer Networks~\cite{TransformersNIPS17}, first to model sequences merely by attention mechanisms. Aside TF, we consider the Bidirectional Transformers BERT~\cite{BERT19}, which forms the basis for the current performer on most NLP tasks~\cite{RoBERTa19}. To the best of our knowledge, this is the first work adopting NLP technique for trajectory forecasting.

\section{The Transformer Model}\label{sec:method}

\begin{figure*}[!t]
\centering
  \includegraphics[width=0.40\textwidth, height = 4.5cm]{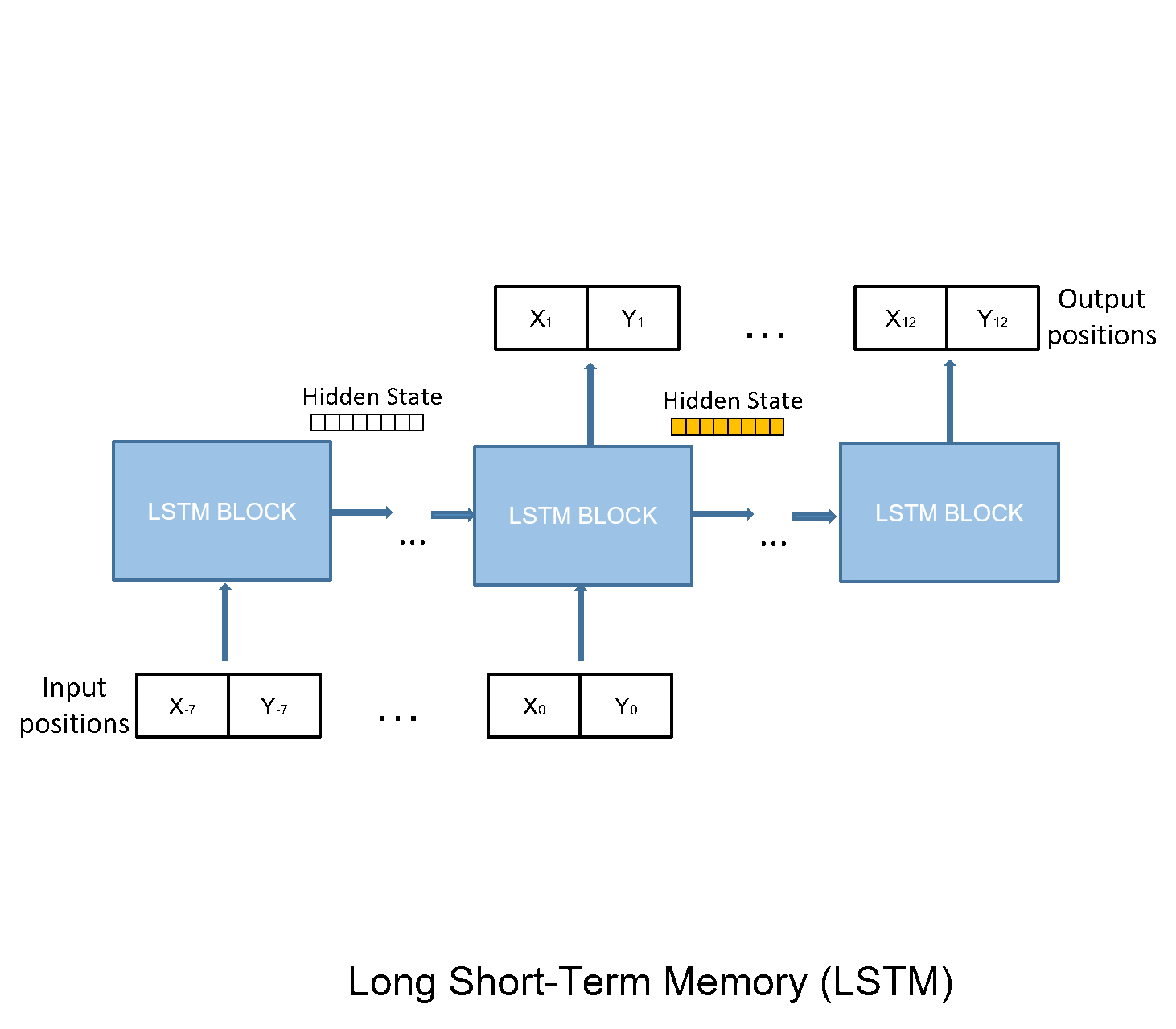}
  \includegraphics[width=0.59 \textwidth, height = 4.5cm]{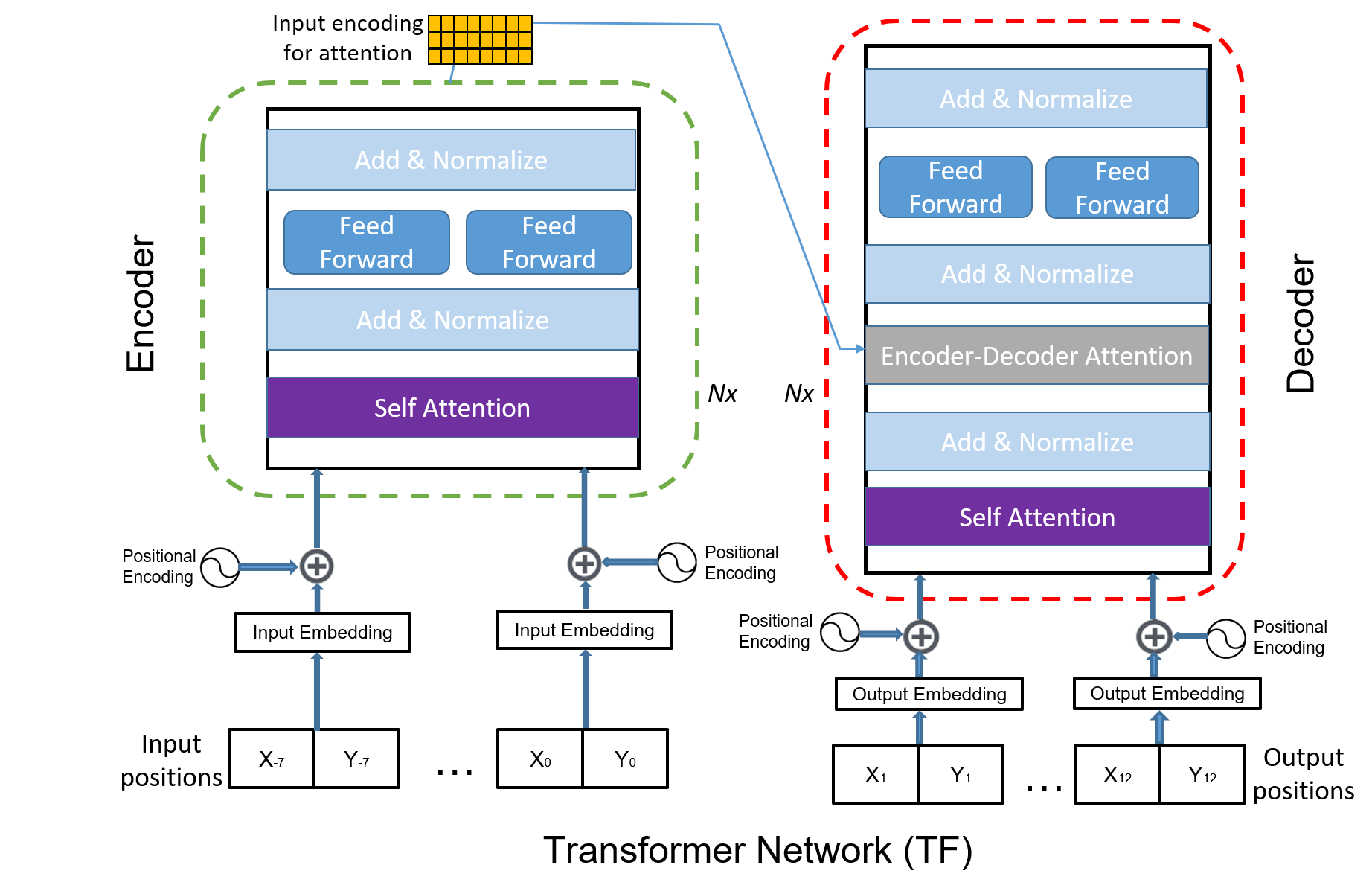}
\vspace{-0.5cm}
\caption{Model illustration of LSTM~(\textit{left}) and TF~(\textit{right}). At each time step, LSTM leverages the current-frame information and its hidden state. By contrast, TF leverages the encoder representation of the observed input positions and the previously predicted outputs. In purple and grey are the self-attention and encoder-decoder attention modules, that allow TF to learn on which past position it needs to focus to predict a correct trajectory.
}
\label{fig:lstm_tf}
%Upon an observation time, when the TF and BERT model of each person ``sees'' the motion history of itsself, they start predicting the future. At each prediction time-step, each person TF and BERT process their own predicted position auto-regressively.
\vspace{-0.2cm}
\end{figure*}

%  \begin{figure*}[t]
% 	\begin{center}
% 	    \includegraphics[width=1\linewidth]{Figures/Model_picture.pdf}
% 	   % \decoRule
% 	\end{center}
% % 	\vspace{-5pt}
% 	\caption{.}\label{fig:prop-appro}
%  	%\vspace{-16pt}
% \end{figure*}

% \begin{figure}[!htb]
% \minipage{0.32\textwidth}
%   \includegraphics[width=\linewidth]{Figures/LSTM.pdf}
%   \caption{}\label{fig:awesome_image1}
% \endminipage\hfill
% \minipage{0.32\textwidth}
%   \includegraphics[width=\linewidth]{Figures/BERT.pdf}
%   \caption{}\label{fig:awesome_image2}
% \endminipage\hfill
% \minipage{0.5\textwidth}%
%   \includegraphics[width=\linewidth]{Figures/tf.pdf}
%   \caption{}\label{fig:awesome_image3}
% \endminipage
% \end{figure}

% \begin{figure}
%     \centering
%     \subfigure[]{\includegraphics[width=0.33\textwidth]{Figures/LSTM.pdf}} 
%     \subfigure[]{\includegraphics[width=0.33\textwidth]{Figures/LSTM.pdf}} 
%     \subfigure[]{\includegraphics[width=0.33\textwidth]{Figures/LSTM.pdf}}
%     \caption{(a) blah (b) blah (c) blah (d) blah}
%     \label{fig:foobar}
% \end{figure}

We propose a multi-agent framework where each person is modelled by an instance of our transformer network. Each Transformer Network predicts the future motion of the person as a result of its previous motion.

We describe in this section the model input and output (Sec.~\ref{sec:inout}), the encoder-decoder Transformer Network (TF) (Sec.~\ref{sec:encdec}) and the just-encoder BERT model (Sec.~\ref{sec:bert}) and the implementation details (Sec.~\ref{sec:modtrain}). %Please refer to Fig.~\ref{fig:lstm_tf} for an overall illustration of our model.

\subsection{Model input and output}\label{sec:inout}

For each person, the transformer network outputs the predicted future positions by processing their current and prior positions (observations or motion history). We detail here each of the input and output information and parallel those with the established LSTM, with reference to Fig.~\ref{fig:lstm_tf}.

% Note: both the TF and BERT models only consider the motion histories of each person separately, without awareness of the social person-person interaction~\cite{some-cites}, nor awareness of the context~\cite{sr-lstm,trajectronpp}.

\paragraph{Observed and predicted trajectories}
In formal terms, for person $i$, we are provided a set $\mathcal{T}_{obs}=\{\mathbf{x}^{(i)}_t\}^0_{t=-(T_{obs}-1)}$ of $T_{obs}$ observed current and prior positions in Cartesian coordinates $\mathbf{x}\in\mathcal{R}^2$, and we are required to predict a set $\mathcal{T}_{pred}=\{\mathbf{x}^{(i)}_t\}^{T_{pred}}_{t=1}$ of $T_{pred}$ predicted positions. %$W_{\mathbf{x}}$
In order to let the transformer deal with the input, this is embedded onto a higher $D$-dimensional space by means of a linear projection %function $\phi$ 
with a matrix of weights $W_{\mathbf{x}}$, i.e., $\mathbf{e}_{obs}^{(i,t)} =  \mathbf{x}^{(i)^{\intercal}}_t  W_{\mathbf{x}}$.

%, followed by a ReLU activation function:
%\begin{equation}
%    \mathbf{e}_{obs}^{(i,t)} = \phi \left( %\mathbf{x}^{(i)}_t,W_{\mathbf{x}} \right)
%\end{equation}
In the same way, the output of our transformer model for person $i$ at time $t$ is the $D$-dimensional vector $\mathbf{e}_{pred}^{(i,t)}$, which is back-projected to the Cartesian person coordinates $\mathbf{x}^{(i)}_t$. LSTM and TF share this aspect.

\paragraph{Positional encoding}
The transformer encodes time for each past and future time instant $t$ with a ``positional encoding''. In other words, each input embedding $\mathbf{e}_{obs}^{(i,t)}$ is time-stamped with its time $t$. The same encoding is used to prompt the model to predict into future instants, as we detail in the next section.

More formally, the input embedding $\mathbf{e}_{obs}^{(i,t)}$ is time-stamped at time $t$ by adding a positional encoding vector $\mathbf{p}^{t}$, of the same dimensionality $D$:
%\begin{equation}\label{eq:posenc}
$
    \mathbf{\xi}_{obs}^{(i,t)} = \mathbf{p}^{t} + \mathbf{e}_{obs}^{(i,t)}
$
%\end{equation}

We use sine/cosine functions to define $\mathbf{p}^{t}$ as in~\cite{TransformersNIPS17}:
%We follow the formulation of \cite{TransformersNIPS17} and use sine/cosine functions to define $\mathbf{p}^{t}$:
\begin{align}
    &\mathbf{p}^{t}=\left\{ p_{t,d} \right\}_{d=1}^{D}\\
    &\text{where}\quad
        p_{t,d}=\left\{
        \begin{tabular}{lr}
        sin$\left(\frac{t}{10000^{d/D}}\right)$ & for $d$ even \\
        cos$\left(\frac{t}{10000^{d/D}}\right)$ & for $d$ odd
        \end{tabular}
    \right.
\end{align}
In other words, each dimension of the positional encoding varies in time according to a sinusoid of different frequency, from $2\pi$ to $10000 \cdot 2\pi$. This ensures a unique time stamp for sequences of up to 10000 elements and extends unseen lengths of sequences. %and allow the model to process input by relative positions, i.e.\ for a given offset $t_o$ the encoding vector $\mathbf{p}^{t+t_o}$ may be represented as a linear function of $\mathbf{p}^{t}$.

In this aspect, TF differs greatly from LSTM, cf.\ Fig.~\ref{fig:lstm_tf}. LSTM processes the input sequentially and the order of input positions determine the flow of time. It does not therefore need a positional encoding. However, LSTM needs to ``unroll'' at training time, i.e.\ back-propagate the signal sequentially across the LSTM blocks processing the observations. By contrast, the training of TF is parallelizable.
%TF may learn from all time instants in parallel, which results in more efficient and scalable training.

Notably, thanks to the positional encoding which time-stamps the input, TF may deal with missing observations.  Missing data is just neglected, but the model is aware of the relative time-stamps of the presented observations. In Sec.~\ref{Sec:exp}, we experiment on this unique feature, important when dealing with real sensor data.

\paragraph{Regression Vs.\ classification}

Regression Vs.\ classification is a recurrent question in trajectory forecasting. Regression techniques, predicting the ($x$,$y$) coordinates directly, generally outperform classification-based approaches, where the inputs are quantized into classes and the input data represented as one-hot-vectors. We test both approaches and confirm the better performance of regression. However, a classification approach, which we dub TF$_q$, provides a probabilistic output across the quantized motions.

We leverage TF$_q$ to sample multiple future predictions, which we assess both quantitatively and qualitative. The predictions of TF$_q$ are multi-modal, as we illustrate in Sec.~\ref{Sec:exp}.

%TF$_q$ outputs softmax'ed probabilities which differ from the Gaussian LSTM~\cite{alahi2016cvpr} probabilistic output. The TF$_q$ outputs are in fact multi-modal, as being generated directly by a deep neural network, while Gaussian LSTM only predicts means and variances of Gaussians, forcing the predictions to a single mode. We illustrate this in Sec.~\ref{Sec:exp}.

% \begin{figure}[t]
% \centering
% \includegraphics[width=\linewidth]{./Figures/EncDecModel.JPEG}
% \caption{Upon an observation time, when the TF and BERT model of each person ``sees'' the motion history of itsself, they start predicting the future. At each prediction time-step, each person TF and BERT process their own predicted position auto-regressively. \todo{Irtiza: please update this figure.}}
% \label{fig:encdecmodel}
% \end{figure}

\subsection{Encoder-decoder Transformer (TF)}\label{sec:encdec}

As illustrated in Fig.~\ref{fig:lstm_tf}, TF is a modular architecture, where both the encoder and the decoder are composed of 6 layers, each containing three building blocks: \textbf{i.}\ an attention module, \textbf{ii.}\ a feed-forward fully-connected module, and \textbf{iii.}\ two residual connections after each of the previous blocks.

The capability of the network to capture sequence non-linearities lies mainly in the attention modules. Within each attention module, an entry of a sequence, named ``query'' (Q), is compared to all other sequence entries, named ``keys'' (K) by a scaled dot product, scaled by the equal query and key $d_k$ embedding dimensionality. The output is then used to weight the same sequence entries, named now ``values'' (V). Attention is therefore given by the equation:
%In practice, each sequence entry is considered as query, and all entries are gathered into matrices of queries \emph{Q}, keys \emph{K} and values \emph{V}, to yield attention according to equation:
%\vspace{-0.3cm}
\begin{equation}\label{eq:att}
    \text{Attention}(Q,K,V)=softmax\left(\frac{Q K^T}{\sqrt{d_k}}\right)
\end{equation}

The goal of the encoding stage is to create a representation for the observation sequence, which makes the model \emph{memory}. To this goal, after the encoding of the $T_{obs}$ input embeddings $\mathbf{\xi}_{s}^{(i,t)}$, the network outputs two vectors of keys $K_{enc}$ and values $V_{enc}$ which would be passed on to the decoder stage.

The decoder predicts auto-regressively the future track positions. At each new prediction step, a new decoder query $Q_{dec}$ is compared against the encoder keys $K_{enc}$ and values $V_{enc}$ according to Eq.~\eqref{eq:att} (encoder-decoder attention) and against the previous decoder prediction (self-attention).
%, to yield the next-step prediction.

Note the important difference w.r.t.\ LSTM: TF maintains the encoding output (memory) separate from the decoded sequence, while LSTM accumulates both into its hidden state, steering what to memorize or forget at each time. We believe this may contribute to explain how TF outperforms LSTM in long-term horizon predictions, cf.\ Sec.~\ref{Sec:exp}.

% \begin{figure}[!ht]
% \centering
%   \includegraphics[width=0.43 \linewidth, height = 4.5cm]{Figures/BERT_.png}
% \vspace{-0.3cm}
% \caption{Model illustration of BERT}\label{fig:bert}
% \end{figure}

\subsection{BERT}\label{sec:bert}

We consider for trajectory forecasting a second Transformer model, BERT~\cite{BERT19}.
Differently from TF, BERT is only composed of an encoder and it trains and infers thanks to a masking mechanism. In other words, the model hides (masks) from the self-attention the output positions which it targets for prediction as the TF decoder also does.
During training the model learns to predict masked positions. At inference, the model output predictions for the masked outputs.

BERT is the \textit{de-facto} reference model for state-of-the-art NLP methods, but larger than TF ($\sim$2.2 times larger). As we would illustrate in Sec.~\ref{Sec:exp}, training BERT on the current largest trajectory forecasting benchmarks does not keep up to the expectations. We draw inspiration from transfer learning and test therefore also how a BERT pre-trained on an NLP task performs on the target task. In particular, we take the lower-cased English text using Whole-Word-Masking; we substitute for the word embedding from dictionary keys with similar linear modules encoding ($x$,$y$) positions; and then we similarly convert also the output into ($x$,$y$) positions.

% At inference time, we apply the same approach, passing as input the observed values and mask tokens for the positions to predict.

\subsection{Implementation details}\label{sec:modtrain}

Our TF implementations adopts the parameters of the original Transformer Networks~\cite{TransformersNIPS17}, namely $d_{model}=512$, 6 layers and 8 attention heads. We adopt an L2-loss between the predicted and annotated pedestrian positions and train the network via backpropagation with the Adam optimizer, linear warm-up phase for the first 5 epoch and a decaying learning rate afterward; dropout value of 0.1.
The normalization of the network input influences its performance, as also maintained in \cite{zhang2019sr,graves2013generating}. So we normalize the people speeds by subtracting the mean and dividing by the standard deviation of the train set.
For the TF$_q$, we quantize the people motion by clustering speeds into 1000 joint ($x$,$y$) bins, then encode the position by 1000-way one-hot vectors. In order to get a good cluster granularity, we augment the training data by random scaling uniformly with scale $s\in[0.5,2]$.

\section{Experimental Evaluation}\label{Sec:exp}

We show the capabilities of the proposed Transformer networks for trajectory forecasting on two recent and large datasets: the TrajNet Challenge~\cite{sadeghiankosaraju2018trajnet} dataset and the ETH+UCY dataset~\cite{pellegrini2009iccv,lerner2007crowds}. Additionally, we perform an ablation study to quantify the model robustness, also in comparison with the widely-adopted LSTM. This includes varying the observation horizon and testing the model on missing data, the latter occurring when some observation samples are missing due to frame-rate drops or excessive uncertainty in the tracking data.

\subsection{The Trajnet Challenge}\label{sec:trajnet}

\noindent\textbf{The TrajNet Dataset:}
At the moment of writing, the TrajNet Challenge\footnote{\url{http://trajnet.stanford.edu/}} ~\cite{sadeghiankosaraju2018trajnet} does represent the largest multi-scenario forecasting benchmark~\cite{rudenko2019human}; the challenge requests to predict 3161 %(343+51+183+85+235+155+630+427+61+12+52+110+35+362+14+334+10+20+30+12)
human trajectories, observing for each trajectory 8 consecutive ground-truth values (3.2 seconds) \emph{i.e.}, $t-7,t-6,\ldots,t$, in world plane coordinates (the so-called \emph{world plane Human-Human} protocol) and fo\-re\-cas\-ting the following 12 (4.8 seconds), \emph{i.e.}, $t+1,\ldots,t+12$.
%The 8-12-value protocol is consistent with the most trajectory forecasting approaches, usually focused on the 5-dataset ETH-univ + ETH-hotel~\cite{pellegrini2009iccv} + UCY-zara01 + UCY-zara02 + UCY-univ~\cite{lerner2007crowds}.
%Trajnet extends substantially the 5-dataset scenario by diversifying the training data, thus stressing the flexibility and generalization one approach has to exhibit when it comes to unseen scenery/situations.
In fact, TrajNet is a superset of diverse datasets that requires to train on four families of trajectories, namely 1) BIWI Hotel~\cite{pellegrini2009iccv} (orthogonal bird's eye flight view, moving people), 2) Crowds UCY \cite{lerner2007crowds} (3 datasets, tilted bird's eye view, camera mounted on building or utility poles, moving people), 3) MOT PETS~\cite{PETS2009} (multi\-sensor, different human activities) and 4) Stanford Drone Dataset~\cite{robicquet2016learning} (8 scenes, high orthogonal bird's eye flight view, different agents as people, cars etc.), for a total of 11448 trajectories. Testing is requested on diverse partitions of BIWI Hotel, Crowds UCY, Stanford Drone Dataset, and is evaluated by a specific server (ground-truth testing data is unavailable for applicants). As a proof of its toughness, it is worth noting that many recent studies restrict on subsets of TrajNet~\cite{deo2020trajectory,haddad2020self,ridel2020scene}, adopting their train/test splits~\cite{van2019safecritic}.  We instead consider the whole TrajNet dataset for our experiments. TrajNet allows to consider concurrent trajectories, so that it is compliant with ``social'' approaches, that can apply. Conversely, it does not allow use raw images, so that approaches which infer on maps as ~\cite{sadeghian2019sophie,ivanovic2019trajectron,salzmann2020trajectron++} cannot apply.   
\vspace{0.2cm}

\noindent\textbf{Metrics:} In agreement with most literature on people trajectory forecasting, the TrajNet performance is measured in terms of: Mean Average Displacement (MAD, equivalently  Average Displacement Error ADE~\cite{ivanovic2019trajectron}), measuring the general fit of the prediction w.r.t. the ground truth, averaging the discrepancy at each time step; Final Average Displacement (FAD, equivalently Final Displacement Error FDE~\cite{ivanovic2019trajectron}), to check the goodness of the prediction at the last time step.
%how near to the ground-truth the prediction will be at the last step.
The average of MAD and FAD is used to rank the approaches.
%The product of MAD and FAD (averaged over the 3161 trajectories) gives a final score which induces a ranking of the approaches.   
\vspace{0.2cm}

\noindent\textbf{Results on TrajNet:}
We report in Table~\ref{Tab:TrajNet} the complete list of 22 \emph{published} comparative approaches, for a total of 39 approaches at the moment of writing; we omit the 18 unpublished results, all of which, apart from one, nonetheless had lower performance than the previous published top-scoring approach REDv3~\cite{becker2018evaluation}.
%This includes the results described in~\cite{becker2018evaluation}, where the authors discuss the performance of several general forecasting algorithms.
In the table, ``Rank'' indicates the absolute ranking over all the approaches, including the unpublished ones; ``Year'' the year of publication of the method; ``Context'' indicates whether the additional social context (the trajectories of the other co-occurring people) is taken into account (``s'') or not (``/'').
% The table merges the results shown in the two TrajNet websites\footnote{\url{http://trajnet.epfl.ch/} and \url{http://trajnet.stanford.edu/}.
% It is worth noting that, at the moment of writing, the EPFL website has been shut down, in favour of the larger TrajNet++ project which has no results so far.}  

The scores in \textcolor{blue}{\textit{blue italic}} refer to the methods proposed in this work (TF, TF$_q$, BERT, BERT\_NLP\_pretrained). 
Surprisingly, the TF model is the new best, with an advantage in terms of both MAD and FAD w.r.t.\ the second REDv3~\cite{becker2018evaluation} and reducing the total error across the 3161 test tracks by $\sim$145 meters. 

It is of interest that the top four approaches (including ours) are individual ones, so no social context is taken into account. These results undoubtedly suggest that in $\sim$3 seconds of individual observation of an individual, much information about his future can be extracted, and TF is the most successful in doing it. 
%much can be predicted for the next 4 seconds (12-steps) by observing 3 seconds (8 steps) of each person independently. 
In fact,  social approaches appear at lower ranks: the first among them is the SR-LSTM~\cite{zhang2019sr}, then the highly-cited Social Forces~\cite{helbing1995social} (rank 9 and 27), the  MX-LSTM~\cite{Hasan18} and Social GAN~\cite{gupta2018social}. 
The quantized TF$_q$ ranks 16th, very probably due to quantization errors.
For the trajnet challenge the TF$_q$ was used in its deterministic mode, i.e.\ the class with highest confidence was selected for the 12 predictions. This is done so because TrajNet is not set up to evaluate best-of-N metric and only a single prediction can be evaluated by the server.

BERT trained from scratch on trajectories ranks 25th; its  NLP-pretrained version, fine-tuned on TrajNet, follows immediately. The BERT performance may indicate that the model does require a way larger amount of training data, which at the present moment is absolutely not comparable to the size of an NLP dataset. For this reason, in the rest of the experiments we will concentrate on the TF. %, more promising at the present moment.

\begin{table}[t]
\begin{center}
\small
\caption{TrajNet Challenge results (world plane Human-Human TrajNet challenge, websites accessed on 26/07/2020). \blu{Blue italic indicates approaches proposed in this work.} \vspace{-0.2cm} \label{Tab:TrajNet}}
\resizebox{1\linewidth}{!}{
\begin{tabular}{cllllccl} \toprule
    \tbf{Rank} & \tbf{Method} & \tbf{Avg} & \tbf{FAD} & \tbf{MAD}  & \tbf{Context}& \tbf{Cit.}& \tbf{Year} \\ \midrule
    %1  & Ikg-TnT & 0.769 & 1.183  & 0.356 & /&\cite{cheng2020amenet}& 2020  \\
    2  & \blu{TF} & \blu{\tbf{0.776}} & \blu{\tbf{1.197}} & \blu{\tbf{0.356}}  & \blu{/} && \blu{2020}  \\
    
    3  & REDv3 & 0.781 & 1.201  & 0.360 & /&\cite{becker2018evaluation}& 2019  \\
%   3  	algortihm_FIFM %	0.7895 %	1.217 %	0.362 
    4  & REDv2 & 0.783 & 1.207  & 0.359   & / &\cite{becker2018evaluation}& 2019 \\
    %5 & physical computation &0.793&	1.226&	0.36\\
    6  & RED & 0.798 & 1.229  & 0.366   & / &\cite{becker2018evaluation}& 2018\\
    7  & SR-LSTM  & 0.816 & 1.261 & 0.37 & s &\cite{zhang2019sr}& 2019\\ 
%     8 & linear_off5 & 0.8185	&1.266	&0.371\\
    9  & S.Forces (EWAP)  & 0.819  & 1.266 & 0.371& s &\cite{helbing1995social}& 1995 \\
 %10 &fishy3&		0.82 &	1.265&	0.375\\
 %11 &lstm_offsets_v4_attention_sa_pa & 0.8185 &	1.256 &	0.381\\
12  & N-Lin. RNN-Enc-MLP &  0.827&  1.276&  0.377 & /&\cite{becker2018evaluation}& 2018\\
13  & N-Lin. RNN &  0.841&  1.300 & 0.381& /&\cite{becker2018evaluation}&  2018 \\
%14 &JHU Offsets MLP5 &	0.844 &	1.304 &	0.384\\
15  & Temp. ConvNet (TCN) & 0.841 &  1.301 & 0.381& /&\cite{bai2018empirical}&  2018 \\
16 & \blu{TF$_{q}$} &	\blu{0.858} & \blu{1.300} & \blu{0.416}&\blu{/}&&  \blu{2020}  \\
17  & N-Linear Seq2Seq &  0.860 & 1.331 & 0.390& /&\cite{becker2018evaluation}&  2018 \\
%18  & Vanilla LSTM (Off)	& 0.887&	1.374&	0.399& /&\cite{LSTM_MATLAB}&  2016 \\
%18 & notSoStupidPredictorSUL	& 0.8865&	1.374&	0.399\\
    18  & MX-LSTM  & 0.887           & 1.374 & 0.399& s &\cite{Hasan18}& 2018\\
%20 &  lstm_offsets_v3	&	0.8885 &	1.37 &	0.407\\
21  & Lin. RNN-Enc.-MLP & 0.892 & 1.381 & 0.404& /&\cite{becker2018evaluation}&  2018 \\
22  & Lin. Interpolation & 0.894 & 1.359 & 0.429& /&\cite{becker2018evaluation}&  2018 \\
%22 & linear simpler & 0.8895 &	1.372 &	0.407\\
24  & Lin. MLP (Off) & 0.896 & 1.384 & 0.407& /&\cite{becker2018evaluation}&  2018 \\
25 & \blu{BERT} &	\blu{0.897}	& \blu{1.354} &	\blu{0.440}& \blu{/}&\cite{BERT19}&  \blu{2020} \\
26 & \blu{BERT\_NLP\_pretrained} &	\blu{0.902} & \blu{1.357} & \blu{0.447}&\blu{/}&&  \blu{2020} \\
    27  & S.Forces (ATTR)  & 0.904  & 1.395 & 0.412& s &\cite{helbing1995social}& 1995 \\
%    25  & StupidPredictorSUL & 0.9035  & 1.395 & 0.412 \\
29  & Lin. Seq2Seq & 0.923 & 1.429 & 0.418& /&\cite{becker2018evaluation}&  2018 \\
30  & Gated TCN & 0.947 & 1.468 & 0.426& /&\cite{bai2018empirical}&  2018 \\
31  & Lin. RNN & 0.951 & 1.482 & 0.420& /&\cite{becker2018evaluation}&  2018 \\
%29  & sgan median	&	1.031 &	1.577 &	0.485 &
32  & Lin. MLP (Pos) & 1.041 & 1.592 & 0.491& /&\cite{becker2018evaluation}&  2018 \\
34 & LSTM &	1.140 & 1.793 & 0.491& /&\cite{LSTM_MATLAB} & 2018 \\
%%MARCO: link at the code used for this LSTM!!!
% 31 & sgan_mean	& 1.0835 &	1.662 &	0.505\\
% 32 & sgan traj dataset &	1.28 &	1.976 &	0.584\\
 36  & S-GAN & 1.334  & 2.107 & 0.561& s &~\cite{gupta2018social}& 2018   \\
% 34 & OSG_v3 &	1.385 &	2.106 &	0.664\\
% 35 &	social lstm_v2 &	1.3865 &	2.098 &	0.675\\
% 36 & social lstm % 	1.5625 %	2.299 %	0.826\\
40  & Gauss. Process  & 1.642           & 1.038 & 2.245& /&\cite{trautman2010iros}&  2010 \\
%38 & SLSTM v1.0	& 2.055 &	3.038 &	1.072\\
42  & N-Linear MLP (Off) & 2.103 & 3.181 & 1.024& /&~\cite{becker2018evaluation}&  2018  \\
% 40 & vanilla_lstm &		2.107 &	3.114 &	1.1\\
% 41 & occupancy_lstm	&	2.1105 &	3.12 &	1.101\\
% 42 & S-LSTM &	2.4185 &	3.569	& 1.268\\
% 43 & social lstm_v3	& 	2.8735 &	4.323 &	1.424 \\
% 44 & lstm_offsets_2	&	7.1775 &	11.718 &	2.637\\
% 45 & InkognitoTracker2 & 17.793 &	22.293 & 13.293 \\
% 46 & InkognitoTracker &	17.8575 &	22.359	& 13.356 \\
% 47 % linear_v02 &	22.907 &	22.856 &	22.958
% 48 % CV &	23.0775	& 22.971 &	23.184\\
%
%
\bottomrule
\end{tabular}
}
\end{center}
\vspace{-0.5cm}

\end{table}

\subsection{The ETH+UCY Benchmark}\label{sec:ethucy}

Prior to TrajNet, most literature have benchmarked forecasting performance on a set of 5 datasets, namely the ETH-univ and ETH-hotel~\cite{pellegrini2009iccv} video sequences and the UCY-zara01, UCY-zara02 and UCY-univ~\cite{lerner2007crowds} videos.

\noindent\textbf{Datasets and metrics:} The ETH+UCY datasets consist overall of 5 videos taken from 4 different scenes (Zara1 and Zara2 are taken from the same camera but at a different time). Following the evaluation protocol of \cite{alahi2016cvpr} we sample from the data each 0.4 seconds to get the trajectories. We observe each pedestrian for 3.2 seconds (8 frames) and get ground-truth data for the next 4.8 seconds (12 frames) to evaluate the predictions. 
The pedestrian positions are converted to world coordinates in meters from the original pixel locations using homography matrices released by the authors.
The evaluation is done with a LOO approach training for 4 dataset and testing on the remaining one.
Recent works brought up some issues with the ETH+UCY dataset, \cite{scholler2020constant} showed that Hotel contains trajectory that go in a different direction than most of the ones in the other 4 dataset, so learning an environmental prior can be difficult without data augmentation like rotation; \cite{zhang2019sr} bring up the issue that ETH is an accelerated video and so by using a sampling rate of 0.4 seconds the trajectory behave in a different way than the ones in the other 4 datasets, they showed how by reducing the sampling rate they were able to improve their results.
We do not take any measure to fix these issues, in order to have a fair comparison against all the other methods that use these dataset using the standard protocol, but during our internal testing we noticed similar improvement when using their sampling rate for ETH.
Performance is evaluated using MAD and FAD, in meters.

\noindent\textbf{Results:} In Table~\ref{Tab:ETHUCY}, we compare on the ETH+UCY against the most recent and best performing approaches: S-GAN~\cite{gupta2018social}, Social-BIGAT~\cite{kosaraju2019social} and Trajectron++~\cite{salzmann2020trajectron++}. Additionally we include the ``individual'' version of S-GAN~\cite{gupta2018social}, which does not leverage the social information. Note in the Table the trend to include and model as much information as possible. The three leading techniques of S-GAN and Trajectron are in fact ``social'', and one of the best performing ones, Social-BIGAT, additionally ingests the semantic map of the environment (``+map''). Additionally, note that best results are obtained by sampling 20 multiple plausible futures and selecting the best one according to best test performance. We dub this here the best-of-20 protocol, which any technique in Table~\ref{Tab:ETHUCY} adopts.

The rightmost column in Table~\ref{Tab:ETHUCY} shows our proposed TF$_q$ model, the only which allow to sample distributions of trajectories. 
%To compare with other techniques, we switch from a regressing TF model (as shown on TrajNet, predicting the $x$ and $y$ future positions) to classifying TF model. In the latter, we quantize the people motion into joint $x$ and $y$ steps by the use of k-means, then encode each person motion with a one-hot vector. We dub the classifying TF model as TF$_q$.
TF$_q$ achieves the second best performance, only 0.10 behind in terms of MAD and 0.10 in terms of FAD. 
%\todo{Marco: please give us your feedback. If you want to show TFq in the TrajNet too, then this explanation needs to go into the model section. Else this stays here, not to raise complains, and we say we introduce it to compare with others on the best-of-20. I'm in favor of this second, and taking it out from the trajnet table.}

Consistently with the TrajNet challenge, an individual forecasting TF$_q$ technique yields a performance surprisingly ahead or comparable with the best social techniques, even if enclosing additional map information. And trend is also reflected by S-GAN~\cite{gupta2018social}, slightly under-performing its individual counterpart.\\

\begin{table}[t]
\begin{center}
\fontsize{9}{10}\selectfont
\caption{Comparison against SoA models following the best-of-20 protocol. The entirety of SoA approaches is rooted on LSTM, and leverages additional information (social, segmented maps). The mere quantized Transformer TF$_q$ is superior to all the social approaches, second only to Trajectron++. %which has access also to maps. 
Actually, only S-GAN-ind~\cite{gupta2018social} and TF$_q$ have the same input and are directly comparable; all of the other performances are reported as reference, written in cursive.} 
\label{Tab:ETHUCY}
%\resizebox{\textwidth}{!}{%
\vspace{-0.5cm}
\resizebox{1\linewidth}{!}{
\begin{tabular}{cccccc} \toprule
     &\multicolumn{4}{c}{LSTM-based}& TF-based \\
     \cmidrule(lr){2-5} \cmidrule(lr){6-6}
     & Individual &\multicolumn{2}{c}{Social} &Soc.+ map&Ind.\\
     \cmidrule(lr){2-2} \cmidrule(lr){3-4} \cmidrule(lr){5-5} \cmidrule(lr){6-6}
    &  S-GAN-ind  & S-GAN      & Trajectron++  & Soc-BIGAT & TF$_q$ \\
    &~\cite{gupta2018social}&~\cite{gupta2018social}&~\cite{salzmann2020trajectron++}&~\cite{kosaraju2019social}&\\
    \midrule
   ETH &0.81/1.52       & \emph{0.87/1.62}      &  \emph{0.35/0.77}   &\emph{0.69/1.29}& \tbf{0.61 / 1.12}                    \\
   Hotel             & 0.72/1.61       & \emph{0.67/1.37}       & \emph{0.18/0.38} & \emph{0.49/1.01} & \tbf{0.18 / 0.30}                     \\
UCY               & 0.60/1.26       & \emph{0.76/1.52}      & \emph{0.22/0.48}  & \emph{0.55/1.32} & \tbf{0.35 / 0.65}                     \\
Zara1              & 0.34/0.69       & \emph{0.35/0.68}       & \emph{0.14/0.28}& \emph{0.30/0.62}  & \tbf{0.22 / 0.38}                     \\
Zara2                & 0.42/0.84       & \emph{0.42/0.84}       & \emph{0.14/0.30}  & \emph{0.36/0.75}& \tbf{0.17 / 0.32}                     \\
\midrule
\emph{Avg}                  & 0.58/1.18       & \emph{0.61/1.21}   & \emph{0.21/0.45}   & \emph{0.48/1.00}   & \tbf{0.31 / 0.55}     \\               
\bottomrule
\end{tabular}
}
\end{center}
\vspace{-0.7cm}

\end{table}

%\todo{You should explain why BERT is not available here.}
Note that the best-of-20 protocol is a sort of upper-bound reachable by sampling-based approaches; therefore, we analyze the behavior of our Transformer-based predictors TF in the single-trajectory deterministic regime as in~\cite{salzmann2020trajectron++}, where each method gives a single prediction. Results are reported in Table~\ref{Tab:ETHUCYdet}.
\begin{table*}[t]
\begin{center}
\fontsize{8}{8}\selectfont
\caption{Comparison against SoA models following the single trajectory deterministic protocol (numbers of other approaches are taken from~\cite{salzmann2020trajectron++} ). Regular font indicates approaches which are comparable with our Transformer-based predictors, since they use a single individual observed trajectory as input. The other approaches have performance in italic, and are displayed as a reference.} 
\label{Tab:ETHUCYdet}
%\resizebox{\textwidth}{!}{%
\resizebox{0.9\linewidth}{!}{
\begin{tabular}{cccccccc} \toprule
     &Linear&\multicolumn{5}{c}{LSTM-based}& {TF-based}\\
     \cmidrule(lr){2-2}\cmidrule(lr){3-7} \cmidrule(lr){8-8}
     &Individual & \multicolumn{2}{c}{Individual} &\multicolumn{2}{c}{Social} &Soc.+ map&Individual\\
      \cmidrule(lr){2-2}\cmidrule(lr){3-4} \cmidrule(lr){5-6} \cmidrule(lr){7-7}\cmidrule(lr){8-8}
     &Interpolat.&LSTM&  S-GAN-ind  & Social      &  Soc.  & Trajectron++& Trasformer\\
     & & \cite{gupta2018social}&  \cite{gupta2018social}  & LSTM~\cite{gupta2018social}      &   Att.~\cite{kosaraju2019social} & \cite{salzmann2020trajectron++} & TF (ours)\\
     \midrule
  ETH &1.33/2.94       & 1.09/2.94      & 1.13/2.21   & \emph{1.09/2.35}&\emph{0.39/3.74}& \emph{0.50/1.19}& \textbf{1.03/2.10}  \\
  Hotel &0.39/0.72       & 0.86/1.91      & 1.01/2.18   & \emph{0.79/1.76}&\emph{0.29/2.64}& \emph{0.24/0.59}& \textbf{0.36/0.71}  \\
 UCY &0.82/1.59       & 0.61/1.31      & 0.60/\textbf{1.28}   & \emph{0.67/1.40}&\emph{0.20/0.52}& \emph{0.36/0.89}& \textbf{0.53}/1.32  \\
  Zara1 &0.62/1.21       & \textbf{0.41/0.88}      & 0.42/0.91   & \emph{0.47/1.00}&\emph{0.30/2.13}& \emph{0.29/0.72}& 0.44/1.00\\
  Zara2 &0.77/1.48       & 0.52/1.11      & 0.52/1.11   & \emph{0.56/1.17}&\emph{0.33/3.92}& \emph{0.27/0.67}& \textbf{0.34/0.76} \\
  \midrule
    avg &0.79/1.59       & 0.70/1.52      & 0.74/1.54   & \emph{0.72/1.54}&\emph{0.30/2.59}& \emph{0.34/0.84}& \textbf{0.54/1.17}  \\
\bottomrule
\end{tabular}
}
\end{center}
\vspace{-0.3cm}

\end{table*}

The message is clear: when it comes to individual approaches, the transformer predictor is better than any individual LSTM-based approach. Notably, TF is better than the Social-LSTM~\cite{gupta2018social}, and it outperforms the Social Attention~\cite{kosaraju2019social} in terms of FAD too, by a large margin. Notably, the only case in which LSTM compares favorably with TF is on Zara1, which is the less structured of the datasets of the benchmark, mostly containing straight lines.

\subsection{Ablation study and qualitative results}\label{sec:abl}

Here we conduct an ablation study on the proposed TF model for forecasting, compare it with the LSTM, and finally illustrate qualitative results.

\subsubsection{Changing the Prediction Lengths \label{Sec:var_length}}

As a first study case, we compare the stability of the TF and LSTM models when predicting longer temporal horizons. Unfortunately, TrajNet does not allow to set the prediction horizon. We set therefore to pursue a test-time experiment of models trained on the large and complex TrajNet on longer video dataset. We collect these from the 5 datasets of ETH+UCY, by selecting those datasets which are not part of the TrajNet training set, namely ETH and Zara01.
In Table~\ref{Tab:var_length}, we vary the observation sequence, from 12 frames (4.8s) to 32 frames (12.8) at a step of 1.8s. %As comparison we evaluate the LSTM~\cite{LSTM_MATLAB}, since it is a common comparison in modern forecasting~\cite{alahi2016cvpr,gupta2018social,sadeghian2019sophie,zhang2019sr,Hasan18,becker2018evaluation}, its code is available (while ranks 2-16 of TrajNet are not), and because it is the basis element of several advanced systems \cite{alahi2016cvpr,Hasan18}. 
Both TF and LSTM have been trained one-dataset-out with training sequences of 8 samples and 12 for the prediction. 

\begin{table}[h] 
\begin{center}
\footnotesize
\caption{MAD and FAD errors when letting the $TF$ and the LSTM models predict longer horizons, i.e.\ from 12 to 32 time steps. Both models were trained on the TrajNet train set, while errors are reported over the union of ETH and Zara1 sequences (not part of the TrajNet train set).}
 \begin{tabular}{ccc}
  \toprule
    {Pred.} & {\textbf{TF (ours)}} & {LSTM~\cite{LSTM_MATLAB}}  \\ 
    {} & {MAD / FAD} & {MAD / FAD}\\\midrule
12      & \textbf{0.71/1.56} & 0.78/1.70 \\
16      & \textbf{0.95/2.15} & 1.15/2.72 \\
20      & \textbf{1.27/2.90} & 1.64/3.99 \\
24      & \textbf{1.66/3.76} & 2.29/5.55 \\
28      & \textbf{2.27/5.09} & 3.07/7.46 \\
32      & \textbf{2.98/4.52} & 4.13/9.96\\
\bottomrule%\\
%\\
%\\
\label{tab:horizon}
\end{tabular}
\end{center}
\label{Tab:var_length}%
\vspace{-0.5cm}

\end{table}
On Table~\ref{tab:horizon} are reported the average MAD and FAD values over the ETH-univ and UCY-zara1. Obviously, performances are generally decreasing. TF has a consistent advantage at every horizon Vs.\ LSTM and the decrease with the horizon of LSTM is approximately 25\% worse, as LSTM degrades from 0.78 to 4.13 MAD, while TF degrades from 0.71 to 2.98 MAD.
\begin{figure*}[!b]
    %\centering
    \includegraphics[width= 1.0\linewidth]{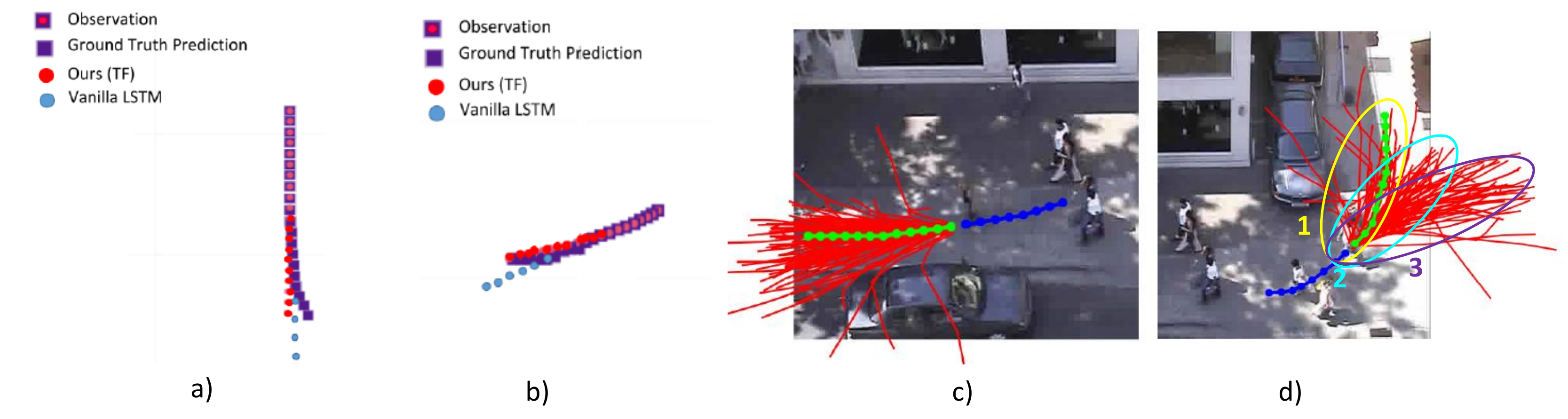}
    \caption{Qualitative results: a) and b) showcasing failures of LSTM, c) and d) illustrating the trajectory distributions learned by TF$_q$. Best viewed in colors.}
\label{fig:trajnet-qualit}
\end{figure*}{}

\subsubsection{Missing and noisy data \label{Sec:missing}}
To the best of our knowledge, the problem of having missing coordinates in \emph{coordinate-based} long-term forecasting\footnote{\emph{Coordinate-based} forecasting takes as input floor coordinates of people, and is different to \emph{image-based} forecasting, where images are processed to extract bounding boxes locations on the image plane such as~\cite{kitani2012activity}.}  has been never taken into account. On the contrary, the problem of missing data is common in short term-forecasting (\emph{i.e.} tracking \cite{bae2017confidence}), or forecasting of heterogeneous data \cite{anava2015online,chen2001study,rodrigues2013spectral,ghazi2018robust,golyandina2007caterpillar}, where in general is treated by designing ad-hoc extensions for filling properly the missing entries (the so called \emph{hindsighting}~\cite{anava2015online}). 
Compared to these techniques, our transformer architecture represents a novel view, since \emph{it does not need to fill missing data}; instead, it exploits the remaining samples knowing when they have been observed thanks to the positional encoding. For example, supposing the $t-k$th sample being missed in the observation sequence, the transformer will use the remaining $t-7,\ldots,t-k-1,t-k+1,\ldots,t$, with $1\leq k \leq8$ to perform the prediction of $t+1,\ldots,t+12$. This structural ability is absent in LSTM and RNN in general (they cannot work with missing data), and in this sense the Transformer is superior. If replacements of missing values can be computed, we found that simple linear interpolation gives slight improvements to the results.

Having witnessed the superiority of the transformer over LSTM in absolute sense (on TrajNet, and see Tab.~\ref{Tab:TrajNet}) and varying the forecasting horizons (Sec.~\ref{Sec:var_length}), we continue this analysis focusing on our proposed model on the same TrajNet dataset. The idea is to systematically drop one element at observation time, at a fixed position, from the most recent (time $t$, indicated also as the \emph{current frame}, after that it starts the prediction) to the furthest ($t-7$).
Results are reported in Tab.~\ref{Tab:MissingNew}.

\begin{table}[t]
\begin{center}
\small
\caption{Evaluation of missing data results for TF on TrajNet. We experiment dropping a varying number of most recent observed samples, either including or excluding the current frame. For example, in the case of dropping 3 frames, we drop $\mathcal{T}_{obs}=\{\mathbf{x}^{(i)}_t\}^0_{t=-2)}$ and  $\mathcal{T}_{obs}=\{\mathbf{x}^{(i)}_t\}^{-1}_{t=-3)}$ respectively.
}
\label{Tab:MissingNew}
\resizebox{1\linewidth}{!}{
\begin{tabular}{ccc} \toprule
    {\# most recent } & {Drop most recent obs.} & {Drop most recent obs.}\\ frames dropped & \emph{including} current frame & \emph{excluding}\ current frame\\ & ({FAD}/{MAD})& ({FAD}/{MAD})\\
    \midrule
0        & 1.197 / 0.356        & 1.197 / 0.356        \\
    \midrule
1        & 1.305/ 0.389         & 1.267 / 0.373        \\
2        & 1.409 / 0.429        & 1.29 / 0.38          \\
3        & 1.602 / 0.495        & 1.303 / 0.384        \\
4        & 1.787 / 0.557        & 1.313 / 0.387        \\
5        & 1.897/ 0.593         & 1.327 / 0.329        \\
6        & 2.128 / 0.669        & 1.377 / 0.406       \\
\bottomrule
\end{tabular}
}
\end{center}
\vspace{-0.5cm}

\end{table}
%\todo{Casino con la spiegazione: piu' sopra si parla di t-7, con t il current frame, sotto in tabella si fa vedere fino a 6...}
The results show that, in a complex scenario such as TrajNet, dropping input frames impact the prediction performance, matching the intuition: the more dropped frames, the larger the performance decrease. Interestingly, the current frame plays a key importance, as it is the most recent observed input, from which future predictions start. In fact, dropping the current frame together with the most recent 6 nearly doubles the error, i.e.\ degrades performance by 91\%, from 0.356 to 0.669 MAD. By contrast dropping 6 observed frames but keeping the current one only degrades the performance by 16\% (from 0.356 to 0.406 MAD), although the TF may now only leverage 2 observations (farther and closest in time).

%\todo{COMPUTATIONAL COMPLEXITY IS MISSING!!!!}

\subsubsection{Qualitative results \label{Sec:qualitative}}
%%MARCO from here
Qualitative results can further motivate the numerical results presented so far. In Fig.~\ref{fig:trajnet-qualit}, we report two predictions assessed on TrajNet, built by using the official visualizer of the benchmark. In particular, we artificially superpose the predicted trajectories of LSTM and TF to highlight their different behavior. In Fig.~\ref{fig:trajnet-qualit} a), the subject is going south, with a minimal acceleration (not immediately visible by the figure, but numerically present); LSTM takes this gentle acceleration, predicting a uniform acceleration toward south. TF captures better the dynamics, despite at the very end the final direction is not correct. 

In Fig.~\ref{fig:trajnet-qualit} b) a similar behavior caused LSTM to predict a faster straight trajectory, while TF followed in this case the bending of the GT more precisely.  

In general, we observed that LSTM generates trajectories way more regular than those predicted by TF, and this is certainly motivated by its unrolling, opposed to the encoder+decoder architecture of TF. This is also the reason why LSTM is so effective on Zara1, consisting essentially in straight trajectories, and so scarce on Hotel (and in general on TrajNet) if compared to TF. 

To further motivate this, in Fig.~\ref{fig:trajnet-qualit} c) and d), we show 100 sampled trajectories by TF$_q$ on Zara1, for two different cases. Fig.~\ref{fig:trajnet-qualit} c)  presents essentially a monomodal distribution, with the samples concentrated around the GT, enriched by few articulated trajectories, that have low probability (they are few), but are still plausible.  Fig.~\ref{fig:trajnet-qualit} d) shows that TF has learnt a multimodal distribution, which has at least three modes, one turning north, the other going diagonal, the third (with larger number of trajectories) going east.

\section{Conclusions}
%\vspace{-0.2cm}
We have proposed the use of Transformers Networks, based on attention mechanisms, to predict people future trajectories. The Transformers, state-of-the-art on all NLP tasks, also perform best on trajectory forecasting. We believe that this questions the widespread use of LSTMs for modelling people motion and that this questions the current formulation of complex social and environmental interactions, which our model does not need for best performance.

In addition to achieving the best performance on people forecasting datasets, the proposed Transfomers have shown better long-term prediction behavior, the capability to predict sensible multiple future trajectories and the unique feature of coping with missing input observations, as it may happen when dealing with real sensor data. Equipped with the better temporal models, we envisage potential to address even larger datasets of long-term sequences, where the importance of social terms may play more crucial roles.

%In this work, we have proposed the first encoder-decoder Transformer network for pedestrian forecasting. This is motivated by the better capabilities of attention-based models, such as the Transformer, to model the non-linearities of people paths, especially emerging from longer observation sequences. This contrasts most recent literature and the current state-of-the-art based on LSTM, which shows plateaued performance.

%Additionally, we have introduced the first multi-agent system based on transformer networks, which we have named the Social Transformer. We have modeled each person with a separate Social Transformer and encoded the social context for each person by means of a social occupancy map. Further to best performance, our social transformer scales better than previous techniques to considering longer and more complex observation inputs. \todo{TBD whether this was demonstrated in the experiments.}  Thus motivated, we plan the future extension of the Social Transformer to also consider the scene semantics. \todo{Review the statement on semantics upon the experiments.}

%These results call for the importance of temporal modelling for progress in the field.

%Leveraging the recent progress of sequence modelling from NLP contexts, we model memory with attention and demonstrate that this better captures the non-linearities of long people trajectories. Our model 

\section{Acknowledgments}
\vspace{-0.1cm}
This work is partially supported by the Italian MIUR through PRIN 2017 - Project Grant 20172BH297: I-MALL - improving the customer experience in stores by intelligent computer vision, and  by  the  project  of  the  Italian  Ministry  of  Education,  Universities  and  Research  (MIUR)  ”Dipartimenti  di  Eccellenza  2018-2022”.

\bibliographystyle{IEEEtran}
\bibliography{main_arxiv}

\begin{thebibliography}{10}
\providecommand{\url}[1]{#1}
\csname url@samestyle\endcsname
\providecommand{\newblock}{\relax}
\providecommand{\bibinfo}[2]{#2}
\providecommand{\BIBentrySTDinterwordspacing}{\spaceskip=0pt\relax}
\providecommand{\BIBentryALTinterwordstretchfactor}{4}
\providecommand{\BIBentryALTinterwordspacing}{\spaceskip=\fontdimen2\font plus
\BIBentryALTinterwordstretchfactor\fontdimen3\font minus
  \fontdimen4\font\relax}
\providecommand{\BIBforeignlanguage}[2]{{%
\expandafter\ifx\csname l@#1\endcsname\relax
\typeout{** WARNING: IEEEtran.bst: No hyphenation pattern has been}%
\typeout{** loaded for the language `#1'. Using the pattern for}%
\typeout{** the default language instead.}%
\else
\language=\csname l@#1\endcsname
\fi
#2}}
\providecommand{\BIBdecl}{\relax}
\BIBdecl

\bibitem{sadeghiankosaraju2018trajnet}
A.~Sadeghian, V.~Kosaraju, A.~Gupta, S.~Savarese, and A.~Alahi, ``Trajnet:
  Towards a benchmark for human trajectory prediction,'' \emph{arXiv preprint},
  2018.

\bibitem{pellegrini2009iccv}
S.~Pellegrini, A.~Ess, K.~Schindler, and L.~Van~Gool, ``You'll never walk
  alone: Modeling social behavior for multi-target tracking,'' in \emph{ICCV},
  2009.

\bibitem{lerner2007crowds}
A.~Lerner, Y.~Chrysanthou, and D.~Lischinski, ``Crowds by example,'' in
  \emph{Computer Graphics Forum}, 2007.

\bibitem{alahi2016cvpr}
A.~Alahi, K.~Goel, V.~Ramanathan, A.~Robicquet, L.~Fei-Fei, and S.~Savarese,
  ``Social {LSTM}: Human trajectory prediction in crowded spaces,'' in
  \emph{CVPR}, 2016.

\bibitem{gupta2018social}
A.~Gupta, J.~Johnson, L.~Fei-Fei, S.~Savarese, and A.~Alahi, ``Social gan:
  Socially acceptable trajectories with generative adversarial networks,'' in
  \emph{IEEE Conference on Computer Vision and Pattern Recognition (CVPR)},
  2018, cONF.

\bibitem{kosaraju2019social}
V.~Kosaraju, A.~Sadeghian, R.~Mart{\'\i}n-Mart{\'\i}n, I.~Reid, H.~Rezatofighi,
  and S.~Savarese, ``Social-bigat: Multimodal trajectory forecasting using
  bicycle-gan and graph attention networks,'' in \emph{Advances in Neural
  Information Processing Systems}, 2019, pp. 137--146.

\bibitem{salzmann2020trajectron++}
T.~Salzmann, B.~Ivanovic, P.~Chakravarty, and M.~Pavone, ``Trajectron++:
  Multi-agent generative trajectory forecasting with heterogeneous data for
  control,'' \emph{arXiv preprint arXiv:2001.03093}, 2020.

\bibitem{ivanovic2019trajectron}
B.~Ivanovic and M.~Pavone, ``The trajectron: Probabilistic multi-agent
  trajectory modeling with dynamic spatiotemporal graphs,'' in
  \emph{Proceedings of the IEEE International Conference on Computer Vision},
  2019, pp. 2375--2384.

\bibitem{sadeghian2019sophie}
A.~Sadeghian, V.~Kosaraju, A.~Sadeghian, N.~Hirose, H.~Rezatofighi, and
  S.~Savarese, ``Sophie: An attentive gan for predicting paths compliant to
  social and physical constraints,'' in \emph{Proceedings of the IEEE
  Conference on Computer Vision and Pattern Recognition}, 2019, pp. 1349--1358.

\bibitem{bai2018empirical}
S.~Bai, J.~Z. Kolter, and V.~Koltun, ``An empirical evaluation of generic
  convolutional and recurrent networks for sequence modeling,'' \emph{arXiv
  preprint arXiv:1803.01271}, 2018.

\bibitem{Luo18}
W.~Luo, B.~Yang, and R.~Urtasun, ``Fast and furious: Real time end-to-end 3d
  detection, tracking and motion forecasting with a single convolutional net,''
  in \emph{Proceedings of the IEEE Conference on Computer Vision and Pattern
  Recognition}, 2018, pp. 3569--3577.

\bibitem{scholler2020constant}
C.~Sch{\"o}ller, V.~Aravantinos, F.~Lay, and A.~Knoll, ``What the constant
  velocity model can teach us about pedestrian motion prediction,'' \emph{IEEE
  Robotics and Automation Letters}, 2020.

\bibitem{becker2018red}
S.~Becker, R.~Hug, W.~Hubner, and M.~Arens, ``Red: A simple but effective
  baseline predictor for the trajnet benchmark,'' in \emph{Proceedings of the
  European Conference on Computer Vision (ECCV)}, 2018, pp. 0--0.

\bibitem{becker2018evaluation}
S.~Becker, R.~Hug, W.~H{\"u}bner, and M.~Arens, ``An evaluation of trajectory
  prediction approaches and notes on the trajnet benchmark,'' \emph{arXiv
  preprint arXiv:1805.07663}, 2018.

\bibitem{TransformersNIPS17}
A.~Vaswani, N.~Shazeer, N.~Parmar, J.~Uszkoreit, L.~Jones, A.~N. Gomez,
  L.~Kaiser, and I.~Polosukhin, ``Transformer attention is all you need,'' in
  \emph{NIPS}, 2017.

\bibitem{BERT19}
J.~Devlin, M.-W. Chang, K.~Lee, and K.~Toutanova, ``Bert pre-training of deep
  bidirectional transformers for language understanding,'' 2019.

\bibitem{SparseTF}
R.~Child, S.~Gray, A.~Radford, and I.~Sutskever, ``Generating long sequences
  with sparse transformers,'' \emph{arXiv preprint:1904.10509}, 2019.

\bibitem{morris2008survey}
B.~T. Morris and M.~M. Trivedi, ``A survey of vision-based trajectory learning
  and analysis for surveillance,'' \emph{IEEE Trans. on Circuits and Systems
  for Video Technology}, vol.~18, no.~8, pp. 1114--1127, 2008.

\bibitem{mccullagh1989generalized}
P.~McCullagh and J.~A. Nelder, ``Generalized linear models, no. 37 in monograph
  on statistics and applied probability,'' 1989.

\bibitem{quinonero2005unifying}
J.~Qui{\~n}onero-Candela and C.~E. Rasmussen, ``A unifying view of sparse
  approximate gaussian process regression,'' \emph{Journal of Machine Learning
  Research}, vol.~6, no.~12, pp. 1939--1959, 2005.

\bibitem{williams1998prediction}
C.~K.~I. Williams, ``Prediction with gaussian processes: From linear regression
  to linear prediction and beyond,'' in \emph{Learning in graphical
  models}.\hskip 1em plus 0.5em minus 0.4em\relax Springer, 1998, pp. 599--621.

\bibitem{priestley1981spectral}
M.~B. Priestley, \emph{Spectral analysis and time series}.\hskip 1em plus 0.5em
  minus 0.4em\relax Academic press, 1981.

\bibitem{akaike1969fitting}
H.~Akaike, ``Fitting autoregressive models for prediction,'' \emph{Annals of
  the institute of Statistical Mathematics}, vol.~21, no.~1, pp. 243--247,
  1969.

\bibitem{hochreiter1997long}
S.~Hochreiter and J.~Schmidhuber, ``Long short-term memory,'' \emph{Neural
  computation}, vol.~9, no.~8, pp. 1735--1780, 1997.

\bibitem{vemula2018social}
A.~Vemula, K.~Muelling, and J.~Oh, ``Social attention: Modeling attention in
  human crowds,'' in \emph{2018 IEEE international Conference on Robotics and
  Automation (ICRA)}.\hskip 1em plus 0.5em minus 0.4em\relax IEEE, 2018, pp.
  1--7.

\bibitem{sadeghian2017tracking}
A.~Sadeghian, A.~Alahi, and S.~Savarese, ``Tracking the untrackable: Learning
  to track multiple cues with long-term dependencies,'' \emph{arXiv preprint
  arXiv:1701.01909}, 2017.

\bibitem{ijcai2017-386}
\BIBentryALTinterwordspacing
H.~Su, J.~Zhu, Y.~Dong, and B.~Zhang, ``Forecast the plausible paths in crowd
  scenes,'' in \emph{Proceedings of the Twenty-Sixth International Joint
  Conference on Artificial Intelligence, {IJCAI-17}}, 2017, pp. 2772--2778.
  [Online]. Available: \url{https://doi.org/10.24963/ijcai.2017/386}
\BIBentrySTDinterwordspacing

\bibitem{su2016crowd}
H.~Su, Y.~Dong, J.~Zhu, H.~Ling, and B.~Zhang, ``Crowd scene understanding with
  coherent recurrent neural networks,'' in \emph{IJCAI}, 2016.

\bibitem{RoBERTa19}
Y.~Liu, M.~Ott, N.~Goyal, J.~Du, M.~Joshi, D.~Chen, O.~Levy, M.~Lewis,
  L.~Zettlemoyer, and V.~Stoyanov, ``Roberta a robustly optimized bert
  pretraining approach,'' arXiv:1907.11692, 2019.

\bibitem{zhang2019sr}
P.~Zhang, W.~Ouyang, P.~Zhang, J.~Xue, and N.~Zheng, ``Sr-lstm: State
  refinement for lstm towards pedestrian trajectory prediction,'' in
  \emph{Proceedings of the IEEE Conference on Computer Vision and Pattern
  Recognition}, 2019, pp. 12\,085--12\,094.

\bibitem{graves2013generating}
A.~Graves, ``Generating sequences with recurrent neural networks,'' \emph{arXiv
  preprint arXiv:1308.0850}, 2013.

\bibitem{rudenko2019human}
A.~Rudenko, L.~Palmieri, M.~Herman, K.~M. Kitani, D.~M. Gavrila, and K.~O.
  Arras, ``Human motion trajectory prediction: A survey,'' \emph{arXiv preprint
  arXiv:1905.06113}, 2019.

\bibitem{PETS2009}
J.~{Ferryman} and A.~{Shahrokni}, ``Pets2009: Dataset and challenge,'' in
  \emph{2009 Twelfth IEEE International Workshop on Performance Evaluation of
  Tracking and Surveillance}, Dec 2009, pp. 1--6.

\bibitem{robicquet2016learning}
A.~Robicquet, A.~Sadeghian, A.~Alahi, and S.~Savarese, ``Learning social
  etiquette: Human trajectory understanding in crowded scenes,'' in
  \emph{European conference on computer vision}.\hskip 1em plus 0.5em minus
  0.4em\relax Springer, 2016, pp. 549--565.

\bibitem{deo2020trajectory}
N.~Deo and M.~M. Trivedi, ``Trajectory forecasts in unknown environments
  conditioned on grid-based plans,'' \emph{arXiv preprint arXiv:2001.00735},
  2020.

\bibitem{haddad2020self}
S.~Haddad and S.-K. Lam, ``Self-growing spatial graph networks for pedestrian
  trajectory prediction,'' in \emph{The IEEE Winter Conference on Applications
  of Computer Vision}, 2020, pp. 1151--1159.

\bibitem{ridel2020scene}
D.~Ridel, N.~Deo, D.~Wolf, and M.~Trivedi, ``Scene compliant trajectory
  forecast with agent-centric spatio-temporal grids,'' \emph{IEEE Robotics and
  Automation Letters}, vol.~5, no.~2, pp. 2816--2823, 2020.

\bibitem{van2019safecritic}
T.~van~der Heiden, N.~S. Nagaraja, C.~Weiss, and E.~Gavves, ``Safecritic:
  Collision-aware trajectory prediction,'' \emph{arXiv preprint
  arXiv:1910.06673}, 2019.

\bibitem{helbing1995social}
D.~Helbing and P.~Molnar, ``Social force model for,'' \emph{Physical review E},
  vol.~51, no.~5, p. 4282, 1995.

\bibitem{Hasan18}
I.~Hasan, F.~Setti, T.~Tsesmelis, A.~Del~Bue, F.~Galasso, and M.~Cristani,
  ``Mx-lstm: mixing tracklets and vislets to jointly forecast trajectories and
  head poses,'' in \emph{CVPR}, 2018.

\bibitem{LSTM_MATLAB}
``{LSTM MATLAB} implementation,''
  \url{https://it.mathworks.com/help/deeplearning/ug/long-short-term-memory-networks.html},
  accessed: 2019-11-08.

\bibitem{trautman2010iros}
P.~Trautman and A.~Krause, ``Unfreezing the robot: Navigation in dense,
  interacting crowds,'' in \emph{IROS}, 2010.

\bibitem{kitani2012activity}
K.~Kitani, B.~Ziebart, J.~Bagnell, and M.~Hebert, ``Activity forecasting,'' in
  \emph{ECCV}, 2012.

\bibitem{bae2017confidence}
S.-H. Bae and K.-J. Yoon, ``Confidence-based data association and
  discriminative deep appearance learning for robust online multi-object
  tracking,'' \emph{IEEE transactions on pattern analysis and machine
  intelligence}, vol.~40, no.~3, pp. 595--610, 2017.

\bibitem{anava2015online}
O.~Anava, E.~Hazan, and A.~Zeevi, ``Online time series prediction with missing
  data,'' in \emph{International Conference on Machine Learning}, 2015, pp.
  2191--2199.

\bibitem{chen2001study}
H.~Chen, S.~Grant-Muller, L.~Mussone, and F.~Montgomery, ``A study of hybrid
  neural network approaches and the effects of missing data on traffic
  forecasting,'' \emph{Neural Computing \& Applications}, vol.~10, no.~3, pp.
  277--286, 2001.

\bibitem{rodrigues2013spectral}
P.~C. Rodrigues and M.~De~Carvalho, ``Spectral modeling of time series with
  missing data,'' \emph{Applied Mathematical Modelling}, vol.~37, no.~7, pp.
  4676--4684, 2013.

\bibitem{ghazi2018robust}
M.~M. Ghazi, M.~Nielsen, A.~Pai, M.~J. Cardoso, M.~Modat, S.~Ourselin, and
  L.~S{\o}rensen, ``Robust training of recurrent neural networks to handle
  missing data for disease progression modeling,'' \emph{arXiv preprint
  arXiv:1808.05500}, 2018.

\bibitem{golyandina2007caterpillar}
N.~Golyandina and E.~Osipov, ``The “caterpillar”-ssa method for analysis of
  time series with missing values,'' \emph{Journal of Statistical planning and
  Inference}, vol. 137, no.~8, pp. 2642--2653, 2007.

\end{thebibliography}

\end{document}